\documentclass[letterpaper]{article}
\usepackage[preprint]{aaai2027}
\usepackage[hyphens]{url}
\usepackage{graphicx}
\usepackage{natbib}
\usepackage{caption}
\usepackage{booktabs}
\usepackage{amsmath,amssymb}
\usepackage{fvextra}
\usepackage{listings}
\title{Where to Decompose: Expert Composition and Structured Prediction for Electrodiagnostic Interpretation}

\author{
    Jongwon Lee\textsuperscript{\rm 1},
    Zhenghong Zhou\textsuperscript{\rm 1},
    Jon Nutt\textsuperscript{\rm 2},
    Eric L. Logigian\textsuperscript{\rm 2},
    Jiebo Luo\textsuperscript{\rm 1}
}
\affiliations{
    \textsuperscript{\rm 1}Department of Computer Science, University of Rochester\\
    \textsuperscript{\rm 2}Department of Neurology, University of Rochester Medical Center\\
    jlee457@ur.rochester.edu
}

\begin{document}

\maketitle

\begin{abstract}
Disease-specific specialization may improve medical interpretation, but simply
combining specialized models does not necessarily improve a strong
domain-adapted model. We study when specialization helps on a cohort of
15{,}247 real electrodiagnostic (EDX) studies. We augment a domain-adapted
Qwen3-8B backbone with 4 selected disease-specific experts, whose claims are
checked against the study measurements before combination. This improves
diagnosis-slot $F_1$ from 0.848 for the backbone alone to 0.883 ($+0.034$,
95\% CI $[+0.021,+0.049]$). Under reference-only $F_1$, which counts
every prediction absent from the physician report as an error, the gain
is $+0.003$ (95\% CI $[-0.013,+0.020]$) and not significant.

Ablations show why each design choice matters. Replacing the backbone with all
6 trained experts reduces $F_1$ to 0.808, and adding unchecked expert claims to
the backbone gives 0.855, no significant gain over the backbone alone. Backbone
preservation contributes $+0.049$ $F_1$ along the ablation path, and claim
checking contributes $+0.028$ when the backbone is kept. Frontier LLMs do not
provide a better fusion mechanism: given both the backbone prediction and
expert claims, GPT-5.6 and Sonnet-4.6 reach only 0.805 and 0.807,
respectively.

We further use the predicted diagnostic slots as an explicit intermediate
state for report generation. Providing these slots improves diagnosis
preservation, the agreement of report diagnoses with the physician reference,
from 0.749 to 0.799 for GPT-5.6 and from 0.604 to 0.758 for Sonnet-4.6. These results support selective expert augmentation of a
domain-adapted backbone and structured diagnosis before report generation.
\end{abstract}

\section{Introduction}

Electrodiagnostic (EDX) interpretation integrates measurements from nerve
conduction studies (NCS), F-wave responses, needle electromyography (EMG), and
neuromuscular ultrasound to identify potentially coexisting diagnoses and
produce a report for physician review. This process is challenging even for
experienced physicians because the measurements are heterogeneous and
diagnostic patterns overlap
\citep{wilbourn1998radiculopathy,dillingham2000lsscreen,
bland2000grading,robinson2008grading}. Interpretations also vary among
physicians \citep{johnsen1995variation} and depend on clinical context
\citep{preston2020emg,katirji2002emg}. Automating EDX interpretation could
therefore reduce reporting burden and improve diagnostic consistency.

Because neuromuscular diseases exhibit distinct diagnostic patterns,
Disease-Specific Experts offer a natural way to decompose the task. However,
it remains unclear whether such decomposition improves upon a single
Domain-Adapted Backbone, whether experts should replace or augment it, and how
their predictions should be combined.

We investigate these questions on a cohort of 15{,}247 real-world EDX studies
using a Domain-Adapted Backbone based on Qwen3-8B \citep{qwen3}. Contrary to
the intuition that specialization should improve performance, simple expert
combinations can perform worse than the backbone alone. The union of six
Disease-Specific Experts achieves diagnosis-slot $F_1=0.808$, compared with
0.848 for the backbone, while unchecked expert augmentation reaches 0.855, not
significantly above the backbone. Frontier LLMs do not resolve the problem:
even when GPT-5.6 \citep{gpt56} and Sonnet-4.6 \citep{sonnet46} receive both
the expert claims and the backbone prediction, they reach only 0.805 and
0.807, respectively.

Our analysis identifies three reasons why expert composition fails. Replacing
the backbone loses diagnoses captured by its full-label-space prediction; not
all experts provide information complementary to the backbone; and expert
claims may be unsupported by the measurements or contain incorrect diagnostic
attributes. We therefore retain the backbone, select complementary experts,
and check their claims against the study measurements before adding diagnoses
absent from the backbone. This Selective Augmentation improves diagnosis-slot
$F_1$ from 0.848 to 0.883 ($+0.034$, 95\% CI $[+0.021,+0.049]$).

We further test whether structured diagnostic predictions improve downstream
report generation. Rather than requiring a generator to infer diagnoses and
verbalize them simultaneously from raw measurements, we provide the predicted
diagnostic slots together with the measurements as an explicit intermediate
state. This separates diagnostic prediction from report generation and
improves diagnosis preservation $F_1$ from 0.749 to 0.799 for GPT-5.6 and
from 0.604 to 0.758 for Sonnet-4.6.

Our contributions are threefold:
\begin{enumerate}

\item \textbf{Systematic analysis of expert composition.}
We show that full expert replacement underperforms a Domain-Adapted Backbone,
that unchecked augmentation does not significantly improve on it, and that
frontier-LLM merging does not recover the gap.

\item \textbf{Selective Augmentation.}
We retain the backbone, check expert claims against the study measurements,
and add verified diagnoses from selected experts only when absent from the
backbone, improving diagnosis-slot $F_1$ from 0.848 to 0.883.

\item \textbf{Structured diagnostic slots for report generation.}
We use predicted diagnostic slots as an explicit intermediate state between
diagnostic prediction and Free-Text Report Generation, improving diagnosis
preservation for both evaluated generators.

\end{enumerate}

\section{Related Work}

\paragraph{EDX interpretation.}
Prior work generated reports directly from EMG tables
\citep{gu2020emgreport}. EMGLLM \citep{long2025emgllm} similarly performs
table-to-report generation, encoding continuous measurements by their degree of
abnormality; its direct-generation setting and clinical judging rubric motivate
our evaluation. Related work has studied clinical-LLM evaluation
\citep{singhal2023medpalm} and structured diagnosis from medical tables
\citep{guo2024dictllm}.

\paragraph{Decomposition and merging.}
Specialized models have been combined through MoE architectures
\citep{shazeer2017moe} and multi-agent clinical reasoning
\citep{tang2024medagents}, sometimes with retrieval augmentation
\citep{lewis2020rag}. We instead study disease-specific experts defined by
diagnostic category and ask whether and how they can improve an already
domain-adapted EDX backbone.

\paragraph{Explicit intermediate states.}
Chain-of-thought prompting \citep{wei2022cot} and scratchpads
\citep{nye2021scratchpad} generate intermediate text before the final output.
Our intermediate representation records each predicted diagnosis in fixed
slots, such as laterality, severity, and localization. These slots are
evaluated independently before report generation.

\section{Method}

\subsection{Task Formulation and Framework Overview}
\label{method:overview}

Let \(X\) denote the raw EDX measurements of a study, comprising NCS, F-wave responses, needle EMG, and neuromuscular ultrasound when available. Patient age, sex, and height are used to select the appropriate normal reference range for each measurement according to laboratory criteria and published reference values. Stage 1 maps \(X\) to a final diagnostic slot set \(S_{\mathrm{final}}\), and Stage 2 maps \(X\) and \(S_{\mathrm{final}}\) to a free-text report \(R\). Each diagnosis is represented as
\begin{equation}
    s = (c, f, l, v, m, r, \rho),
    \label{eq:diagnostic_slot}
\end{equation}
where \(c\), \(f\), \(l\), \(v\), \(m\), \(r\), and \(\rho\) denote the canonical diagnosis, fiber type, side, severity, mechanism, anatomical region, and involved roots, respectively. A study may contain multiple diagnostic slots, whereas a normal study is represented by the empty set \(\varnothing\). Findings supported only by clinical history, symptoms, physical examination, or external imaging are stored separately and excluded from structured-diagnosis evaluation because these sources are unavailable to the diagnostic models.

\begin{figure*}[t]
    \centering
    \includegraphics[width=\textwidth]{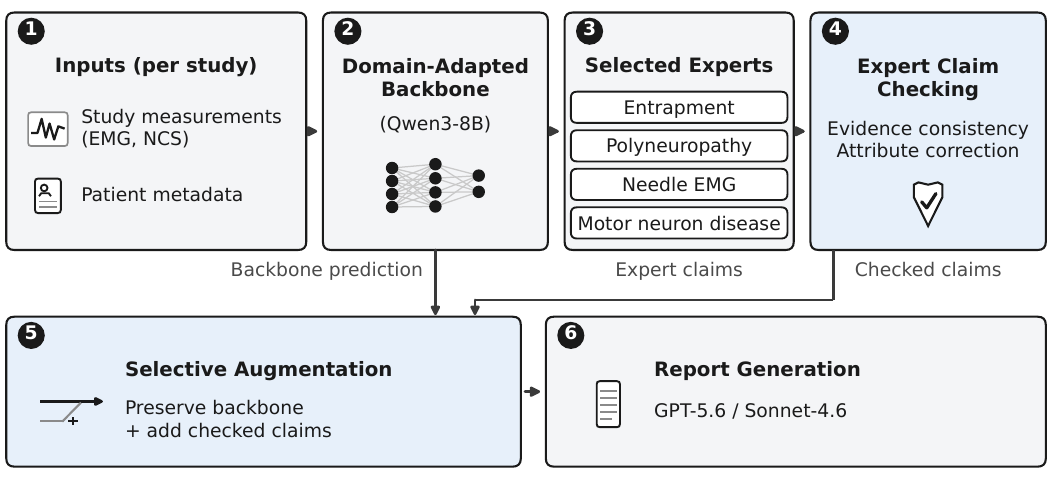}
   \caption{%
Overview of the proposed framework. Four selected experts propose diagnoses
conditioned on the measurements and backbone prediction. Checked expert claims
are selectively added to the preserved backbone prediction, and the resulting
diagnostic slots guide free-text report generation.%
}
    \label{fig:pipeline}
\end{figure*}

Our framework addresses three challenges: preserving diagnostic coverage across the full label space, selecting Disease-Specific Experts that complement the Domain-Adapted Backbone, and filtering unsupported expert claims. As illustrated in Figure~\ref{fig:pipeline}, the Domain-Adapted Backbone first maps \(X\) to a first-pass diagnostic slot set \(S_B\). Each Disease-Specific Expert then receives \(X\) and \(S_B\) and produces candidate diagnostic slots together with a summary of the relevant measurements and their deviations from normal. Selective Augmentation checks the candidate diagnoses against the study measurements and applies a Backbone-Preserving Additive Merge to obtain \(S_{\mathrm{final}}\), retaining the backbone prediction while adding verified expert diagnoses absent from \(S_B\). Finally, Free-Text Report Generation conditions on \(X\) and \(S_{\mathrm{final}}\) to produce a draft report for physician review.

\subsection{Domain-Adapted Backbone}
\label{method:backbone}

Because EDX studies may contain multiple coexisting diagnoses supported by overlapping measurements, a unified interpretation across diagnostic categories is necessary. We use Qwen3-8B~\citep{qwen3} as the Domain-Adapted Backbone and adapt it through supervised fine-tuning with Quantized Low-Rank Adaptation (QLoRA)~\citep{dettmers2023qlora}. The base weights are quantized to 4-bit NF4 and frozen, while a low-rank adapter~\citep{hu2022lora} is trained on the full-label-space diagnostic task. Clinical decision rules and numerical cutoffs are included in the prompt. All Stage-1 models use the same frozen Qwen3-8B base and QLoRA configuration but have independently trained adapters; inference uses greedy decoding. Full training and inference details are provided in Supplements~S5 and~S6.

Given the raw EDX measurements \(X\), the Domain-Adapted Backbone predicts
\begin{equation}
    S_B = B(X),
    \label{eq:backbone_prediction}
\end{equation}
where \(S_B\) denotes the first-pass diagnostic slot set over the full label space. The backbone provides the diagnostic basis for subsequent augmentation but may still miss disease-specific findings, motivating the Disease-Specific Experts introduced next.

\subsection{Disease-Specific Experts}
\label{method:expert}

Each Disease-Specific Expert is an independent QLoRA adapter over the same frozen Qwen3-8B base. For domain \(d\), the expert receives the raw EDX measurements \(X\) and the backbone prediction \(S_B\), and outputs both candidate diagnostic slots and a measurement summary:
\begin{equation}
    (C_d, M_d) = E_d(X, S_B),
    \label{eq:expert_prediction}
\end{equation}
where \(C_d\) denotes the set of candidate diagnostic slots proposed within domain \(d\), and \(M_d\) contains the relevant raw measurements and their deviations from the corresponding normal ranges. The measurement summary records the information used by the expert to support its decisions but is independently checked against \(X\) during Expert Claim Checking.

Conditioning on \(S_B\) allows each expert to focus on diagnoses potentially missed by the backbone rather than regenerating the complete diagnostic slot set.

\subsection{Selective Augmentation}
\label{method:select}
Selective Augmentation applies Expert Claim Checking followed by a Backbone-Preserving Additive Merge. Let \(\mathcal{D}^{*}\) denote the selected expert set. We collect the paired outputs of the selected experts as
\begin{equation}
    \mathcal{O}
    =
    \left\{
        (C_d, M_d)
        \mid
        d \in \mathcal{D}^{*}
    \right\}.
    \label{eq:selected_expert_outputs}
\end{equation}
Retaining the pairing between \(C_d\) and \(M_d\) allows each expert's diagnostic claims to be checked against the measurements and deviations reported by that expert.

Expert Claim Checking applies three deterministic operations to \(\mathcal{O}\) using the raw EDX measurements \(X\):
\begin{enumerate}
    \item \textbf{Internal consistency.}
    For each candidate diagnostic slot \(s \in C_d\), the checker verifies that the measurements and deviations reported in \(M_d\) agree with \(X\) and support \(s\). The candidate is removed otherwise.

    \item \textbf{Sub-label correction.}
    A supported diagnosis is retained, but attributes such as laterality and severity are corrected according to the location and magnitude of the measured abnormalities.

    \item \textbf{Attribution.}
    The same abnormality is not attributed to multiple diagnoses. For example, sensorimotor polyneuropathy is reduced to sensory polyneuropathy when the motor abnormality is explained by another disease.
\end{enumerate}

Together, these operations produce
\begin{equation}
    \widetilde{S}_E = V(\mathcal{O}, X),
    \label{eq:verified_expert_slots}
\end{equation}
where \(V\) retains or corrects supported claims and removes unsupported ones;
\(\widetilde{S}_E\) therefore contains only verified expert slots.

The Backbone-Preserving Additive Merge then implements an add-only rule over \(S_B\) and \(\widetilde{S}_E\). Let \(\kappa(s)\) denote the canonical diagnosis of slot \(s\), and define
\[
    \kappa(S) = \{\kappa(s) \mid s \in S\}.
\]
The final diagnostic slot set is
\begin{equation}
    S_{\mathrm{final}}
    =
    S_B
    \cup
    \left\{
        s \in \widetilde{S}_E
        \mid
        \kappa(s) \notin \kappa(S_B)
    \right\}.
    \label{eq:additive_merge}
\end{equation}
Thus, every backbone diagnosis is preserved, and a verified expert diagnosis is added only when its canonical diagnosis is absent from \(S_B\). The Disease-Specific Experts cannot delete, modify, or overwrite existing backbone diagnoses. The resulting \(S_{\mathrm{final}}\) is passed to Free-Text Report Generation.

\subsection{Free-Text Report Generation}
\label{method:report}

The Free-Text Report Generation module receives the raw EDX measurements \(X\) and the final diagnostic slot set \(S_{\mathrm{final}}\):
\begin{equation}
    R = G(X, S_{\mathrm{final}}),
    \label{eq:report_generation}
\end{equation}
where \(R\) denotes the generated report. The diagnostic slots specify which diagnoses and attributes should be reported, while the generator uses \(X\) to organize the predicted diagnoses and their relevant measurements into clinically readable text. The expert-provided measurement summaries are used only during Expert Claim Checking and are not passed separately to the report generator. This design separates diagnostic prediction from language generation and is intended to reduce omissions, unsupported additions, and attribute changes. At inference time, only predicted diagnostic slots are used; physician-derived reference slots are not required. The output is intended as a draft for physician review.

\section{Experiments}

\subsection{Experimental setup}
\label{sec:metrics}
\label{sec:judge}
\label{sec:stage2eval}

\paragraph{Dataset.}
The cohort comprises 15{,}247 consecutively collected EDX studies, each
providing the measurements $X$ defined above. Normal reference ranges follow
our laboratory's criteria and published reference values
\citep{chen2016edxrefvalues,dillingham2016ndtf}. The reference diagnosis for a
study is the set of slots of the form of Eq.~\eqref{eq:diagnostic_slot},
derived from the attending physician's final report.

\paragraph{Splits.}
We hold out two disjoint sets from the cohort. The primary evaluation set
(eval404, $n=404$) preserves the source cohort's proportion of normal and
abnormal studies; carpal tunnel syndrome (CTS) is the most common diagnosis
(39.4\%), followed by
radiculopathy, ulnar neuropathy at the elbow, and polyneuropathy
(Supplement~S2). The second set contains 698 studies, of which 694 have
backbone predictions and form the expert-selection set used for subset
ablation. This set also covers four diagnoses absent from eval404, including
neuromuscular junction (NMJ) disorder; its composition and diagnostic coverage are detailed in
Supplement~S9.

The backbone is trained on 13{,}140 of the remaining studies, with 703 for
validation. Expert supervision is domain-specific: the 4 retained experts are
trained on 8{,}335 (entrapment), 3{,}919 (polyneuropathy), 6{,}144 (needle
EMG), and 774 (motor neuron disease, MND) examples from the training split (Supplementary
Table~S5). All training, selection, and evaluation partitions are
patient-disjoint.

\paragraph{Expert-domain selection.}
The 4 expert domains used in the reported system were selected on the
694-study set by scoring all 63 non-empty subsets of the 6 experts and taking
the highest-scoring subset. The selected 4-expert configuration was then
evaluated on eval404, which was reserved for the main comparison. The full
subset ablation is reported in Section~\ref{sec:ablations}.

\paragraph{Implementation.}
Frontier models are queried at temperature 0. Full training, serving, and
prompt configurations for the backbone and the 6 experts are in Supplement~S5.

\paragraph{Compared methods.}
We compare off-the-shelf models on the raw measurements, the domain-adapted
backbone alone, expert-only systems, and frontier LLMs that merge expert
claims with or without the backbone prediction. We compare these with our
system, which retains the backbone and adds checked claims from 4 expert
domains.

\subsubsection{Stage-1 evaluation}

\paragraph{Diagnosis-slot $F_1$.}
Predicted and reference slots are greedily matched one-to-one within each
canonical diagnosis. Matched diagnoses receive partial credit for laterality and root-level
agreement \citep{wilbourn1998radiculopathy,dillingham2000lsscreen}.
We generally do not penalize severity disagreement because severity grading can
depend on the grading scheme used; this variability is well documented even
for common conditions such as CTS
\citep{bland2000grading,robinson2008grading}.
The weights were set with our physicians before any system was scored
(Table~S15).

Because the model receives study measurements but not clinical history,
physician reference diagnoses may reflect information unavailable to the
model. We therefore distinguish measurement-supported disagreements from
errors. An unmatched prediction is classified as a measurement-supported
disagreement (DIS) if supported by the study, an ordinary false positive (FP)
if unsupported, or an aggravated FP with $1.5\times$ weight if the reference
explicitly rules it out.

Our primary $F_1$ does not penalize DIS and gives partial credit for attribute
agreement as described above. \emph{Reference-only} $F_1$ instead treats every
prediction absent from the physician reference as an error and gives full
credit to matched diagnoses. Precision and recall are computed from the same
partial-credit matches, with unmatched reference slots counted as false
negatives (FNs). They therefore do not correspond directly to the raw
TP/FP/FN/DIS counts in Table~\ref{tab:errors}. Evidence and subtype-matching
rules are given in Supplement~S10, with the DIS analysis in Supplement~S4.

\paragraph{Laterality normalization.}
When one system reports a diagnosis as bilateral and the other reports the
same diagnosis by side, we split the bilateral prediction into left and right
slots before matching. Results without normalization are reported in
Supplement~S1.

\paragraph{Clinical judging of Stage-1 predictions.}
Our two $F_1$ measures depend on our scoring rules for disagreements and
attribute mismatches. We therefore evaluate the same predictions with a
separate rubric that does not use these rules. Each system's diagnostic slots
are converted into short clinical sentences using a fixed dictionary without
adding diagnostic information.

Two blinded judges, Gemini-3.6-flash \citep{gemini36flash} and
DeepSeek-v4-flash \citep{deepseekv4flash}, compare these verbalized
predictions with the attending physician's report. Following
\citet{long2025emgllm}, correctness counts false or unsupported abnormalities
and completeness counts missed ones, each on a five-point scale. We
additionally count hallucinated and omitted diagnoses per report. Judging is
at temperature 0, with presentation order randomized. The verbatim prompt
is in Supplement~S11.

\subsubsection{Stage-2 evaluation}

\paragraph{Conditions.}
Stage 2 evaluates the reports generated under conditions A--D: \textbf{A}, raw
measurements alone; \textbf{B}, raw measurements with reference slots;
\textbf{C}, raw measurements with predicted slots; and \textbf{D}, predicted
slots alone. We use two complementary measures.

\paragraph{Diagnosis-preservation metric.}
We extract diagnostic slots from each generated report with our report parser
and compare them with the reference slots. Precision, recall, and $F_1$
therefore measure how accurately the reference diagnoses are represented in
the generated report. We refer to this $F_1$ as \emph{diagnosis preservation}.

\paragraph{Clinical judging of generated reports.}
The same two blinded judges, Gemini-3.6-flash and DeepSeek-v4-flash, also rate
the generated reports directly, on correctness and completeness and under the
same blinding. Unlike the Stage-1 judging, where the text is a fixed
verbalization of slots, here the judged text is the free-text report the
generator wrote.

\subsubsection{Statistical analysis}
We compute 95\% percentile CIs using a paired case-level bootstrap with
10{,}000 resamples and a fixed seed; intervals excluding zero are considered
significant.

\subsection{Main results on structured diagnosis}

\begin{table*}[t]
\centering
\small
\setlength{\tabcolsep}{5pt}
\begin{tabular}{@{}lccccl@{}}
\toprule
System & $F_1$ & ref.-only & Precision & Recall & $\Delta F_1$ (system $-$ ours) [95\% CI] \\
\midrule
\multicolumn{6}{@{}l}{\textit{Off-the-shelf models reading the raw measurements}} \\
Qwen3-8B, raw tables   & 0.4911 & 0.4781 & 0.5900 & 0.4210 & $-0.3916$ [$-0.432$, $-0.351$] \\
Sonnet-4.6, raw tables & 0.5921 & 0.5943 & 0.6000 & 0.5840 & $-0.2906$ [$-0.333$, $-0.249$] \\
GPT-5.6, raw tables    & 0.7392 & 0.7404 & \textbf{0.9250} & 0.6160 & $-0.1435$ [$-0.181$, $-0.107$] \\
\midrule
\multicolumn{6}{@{}l}{\textit{Local 8B systems}} \\
Expert union, no backbone & 0.8080 & 0.7603 & 0.8922 & 0.7383 & $-0.0747$ [$-0.099$, $-0.051$] \\
Backbone alone & 0.8484 & 0.8134 & 0.9245 & 0.7838 & $-0.0343$ [$-0.049$, $-0.021$] \\
Backbone $+$ checked claims (ours) & \textbf{0.8827} & \textbf{0.8166} & 0.9166 & \textbf{0.8511} & --- (reference) \\
\bottomrule
\end{tabular}
\caption{Diagnosis-slot accuracy on the 404-case held-out set. \textit{ref.-only}
is reference-only $F_1$; both metrics are defined in Section~\ref{sec:metrics}.
Our system admits the 4 retained expert domains. Differences are paired
bootstrap estimates against ours over 10{,}000 case resamples with a fixed seed;
every interval excludes zero. Laterality is normalized so that a bilateral label
and the corresponding left/right pair are treated as equivalent. Frontier LLMs
used as mergers of expert claims are reported separately in
Table~\ref{tab:arbiter}.}
\label{tab:main}
\end{table*}

Table~\ref{tab:main} shows three main patterns.

First, selective expert augmentation improves structured diagnosis over the
domain-adapted backbone. The backbone alone reaches $F_1=0.848$, whereas
retaining it and adding checked claims from the 4 expert domains selected on
the 694-study expert-selection set reaches 0.883, a gain of $+0.034$ (95\% CI
[$+0.021$, $+0.049$]). Recall increases from 0.784 to 0.851, while precision
decreases slightly from 0.925 to 0.917. The gain therefore comes primarily
from recovering diagnoses missed by the backbone
(Table~\ref{tab:errors}).

Second, experts are less effective when they replace the backbone. The union
of all 6 experts scores 0.808, below the backbone's 0.848. The broader
comparison also shows the importance of domain adaptation: the fine-tuned
Qwen3-8B backbone outperforms every off-the-shelf configuration we tested.
Merging expert claims improves GPT-5.6 over raw-measurement inference by 0.053,
with retrieval adding a further 0.011, but neither closes the gap to the
domain-adapted backbone.

Third, the improvement is not uniform across scoring rules. Under
reference-only $F_1$, the backbone scores 0.813 and our system 0.817
($+0.003$, 95\% CI [$-0.013$, $+0.020]$), so the gain is not significant
under reference-only $F_1$. DIS increase from 50 to 89, and
reference-only $F_1$ treats these measurement-supported diagnoses absent from
the physician reference as errors. Table~\ref{tab:errors} further separates
missed diagnoses, unsupported false positives, and measurement-supported
disagreements.

\paragraph{LLM-based evaluation of Stage-1 predictions.}
Two independent LLM judges confirm the recall--precision trade-off:
relative to the backbone, our predictions are more complete and have fewer
omissions, but are less conservative and contain more additional diagnoses
(Supplementary Table~S2). LLM-merger outputs have substantially lower
completeness than ours without significantly higher correctness. Full scores,
paired comparisons, and judge agreement are reported in Supplementary
Tables~S2 and S4.

\subsection{Diagnosis-grounded report generation}

We next evaluate whether the diagnostic slots produced in Stage 1 improve
free-text report generation. Using GPT-5.6 and Sonnet-4.6 as generators, we
compare the four input conditions of Section~\ref{sec:stage2eval}.

\paragraph{Diagnosis preservation.}
Adding predicted slots improves diagnosis preservation for both generators:
GPT-5.6 increases from 0.749 to 0.799 ($+0.051$, 95\% CI
$[+0.018,+0.083]$), and Sonnet-4.6 from 0.604 to 0.758
($+0.154$, $[+0.114,+0.195]$; Table~\ref{tab:stage2}). Reference slots provide
a further upper-bound comparison. On the 103 normal studies, adding predicted
slots lowers the rate at which GPT-5.6 calls the study normal from 94.2\%
to 88.3\% (Supplementary Table~S7).

\paragraph{LLM evaluation of generated reports.}
For Sonnet-4.6, the LLM judges rate predicted-slot conditioning higher in both
correctness and completeness. For GPT-5.6, neither changes significantly
(Supplementary Table~S3 and Supplement~S7).

\begin{table*}[t]
\centering
\small
\begin{tabular}{@{}lcccccc@{}}
\toprule
& \multicolumn{3}{c}{GPT-5.6} & \multicolumn{3}{c}{Sonnet-4.6} \\
\cmidrule(lr){2-4}\cmidrule(lr){5-7}
Input condition & $F_1$ & P & R & $F_1$ & P & R \\
\midrule
A \ Raw measurements only   & 0.749 & 0.897 & 0.642 & 0.604 & 0.670 & 0.550 \\
B \ Raw $+$ reference slots & \textbf{0.906} & 0.976 & 0.845 & \textbf{0.881} & 0.933 & 0.834 \\
C \ Raw $+$ predicted slots & 0.799 & 0.896 & 0.722 & 0.758 & 0.849 & 0.684 \\
D \ Predicted slots only    & 0.719 & 0.898 & 0.600 & 0.767 & 0.888 & 0.674 \\
\midrule
\multicolumn{7}{@{}l}{\textit{Paired differences in $F_1$ (10{,}000 resamples, fixed seed)}} \\
C $-$ A \ Adding predicted slots
& \multicolumn{3}{l}{$+0.051$ \, [$+0.018$, $+0.083$]}
& \multicolumn{3}{l}{$+0.154$ \, [$+0.114$, $+0.195$]} \\
C $-$ D \ Adding raw measurements
& \multicolumn{3}{l}{$+0.081$ \, [$+0.053$, $+0.110$]}
& \multicolumn{3}{l}{$-0.009$ (n.s.)} \\
\bottomrule
\end{tabular}
\caption{Diagnosis preservation in Stage-2 report generation. The generator and
prompt are fixed; only the input changes. Generated reports are parsed into
diagnostic slots and compared with the reference. Condition~B uses
physician-derived reference slots and serves as a comparison.}
\label{tab:stage2}
\end{table*}

\subsection{Ablations and additional analysis}
\label{sec:ablations}
\paragraph{Backbone preservation.}
We first test whether the experts should replace or augment the
domain-adapted backbone. Table~\ref{tab:ablation} separates this choice from
claim checking and expert selection.
\begin{table*}[t]
\centering
\begin{tabular}{@{}lcccl@{}}
\toprule
Configuration & $F_1$ & P & R & $\Delta F_1$ vs.\ backbone [95\% CI] \\
\midrule
\multicolumn{5}{@{}l}{\textit{Experts replace the backbone}} \\
Expert union$^{\dagger}$
    & 0.8080 & 0.892 & 0.738
    & $-0.0404$ \, [$-0.070$, $-0.011$] \\
Expert union $+$ checking
    & 0.8285 & 0.914 & 0.758
    & $-0.0198$ \, [$-0.046$, $+0.006$] \\
\midrule
\multicolumn{5}{@{}l}{\textit{Experts are added to the backbone}} \\
Backbone alone
    & 0.8484 & 0.925 & 0.784
    & --- \\
Backbone $+$ unchecked claims, 4 domains
    & 0.8545 & 0.845 & 0.864
    & $+0.0061$ \, [$-0.011$, $+0.024$] \\
Backbone $+$ checked claims, 6 domains
    & 0.8773 & 0.905 & 0.851
    & $+0.0289$ \, [$+0.014$, $+0.044$] \\
Backbone $+$ checked claims, 4 domains (ours)
    & \textbf{0.8827} & 0.917 & 0.851
    & $+0.0343$ \, [$+0.021$, $+0.049$] \\
\bottomrule
\end{tabular}
\caption{Stage-1 ablation of expert composition. Intervals are paired-bootstrap
95\% CIs from 10{,}000 case resamples. $^{\dagger}$Simple union of all 6
experts.}
\label{tab:ablation}
\end{table*}

Replacing the backbone with all 6 experts reduces $F_1$ from 0.848 to 0.808.
Checking raises the expert-only system to 0.829, which does not exceed the
backbone ($-0.020$, 95\% CI [$-0.046$, $+0.006$]), whereas retaining the
backbone and adding the same checked claims reaches 0.877
(Table~\ref{tab:ablation}). Along the path from unchecked expert union to our
final system, checking, backbone preservation, and 4-domain selection
contribute $+0.021$ [$+0.004$, $+0.039$], $+0.049$ [$+0.029$, $+0.068$], and
$+0.005$ [$+0.002$, $+0.010$] $F_1$, respectively. Thus, most of the gain comes
from preserving the backbone and checking expert claims.

\paragraph{Evidence verification.}
With the same 4 domains, checking expert claims raises $F_1$ from 0.855 to
0.883 ($+0.028$, 95\% CI [$+0.019$, $+0.038$]). Precision rises from 0.845 to
0.917, and recall falls from 0.864 to 0.851 (Table~\ref{tab:ablation}).

\paragraph{Specialist selection.}
We select the expert domains on the separate 694-study expert-selection set
before evaluating the resulting configuration on eval404. We evaluate all 63
non-empty subsets of the 6 experts on this selection set. The combination of
entrapment, polyneuropathy, needle-EMG, and MND experts achieves the highest
$F_1$ across all subset sizes, reaching 0.8143 compared with 0.7820 for the
backbone. Admitting all 6 experts scores 0.8078 ($-0.0065$, 95\% CI
$[-0.0105,-0.0030]$).

The margins at the top are small. The best 3-expert subset, the same set
without MND, reaches 0.8135, 0.0008 below the selected set (95\% CI
$[-0.0024, 0.0000]$), and the best 5-expert subset, which adds NMJ, reaches
0.8132. The best 2-expert subset reaches 0.8050 ($-0.0092$,
$[-0.0146,-0.0045]$), and the best single expert, entrapment, reaches 0.7948.
Supplement~S9 gives the full ranking.

\paragraph{Fusion strategy.}
We replace the deterministic merge with GPT-5.6 or Sonnet-4.6 while holding
the available predictions and evaluation protocol fixed (Table~\ref{tab:arbiter}).
\begin{table*}[t]
\centering
\small
\begin{tabular}{@{}lccccl@{}}
\toprule
System & $F_1$ & \multicolumn{3}{c}{Policy compliance} & $\Delta F_1$ (system $-$ ours) [95\% CI] \\
\cmidrule(lr){3-5}
 & & drop & lat. & admit & \\
\midrule
GPT-5.6 merger & 0.7919 & --- & --- & --- & $-0.091$ \, [$-0.119$, $-0.062$] \\
\quad $+$ retrieval & 0.8026 & --- & --- & --- & $-0.080$ \, [$-0.107$, $-0.053$] \\
\quad $+$ backbone & 0.8045 & 0.069 & 78.5\% & 22.0\% & $-0.078$ \, [$-0.103$, $-0.052$] \\
\quad $+$ backbone, add-only & 0.8057 & 0.002 & 100\% & 46.6\% & $-0.077$ \, [$-0.097$, $-0.057$] \\
\midrule
Sonnet-4.6 merger & 0.7320 & --- & --- & --- & $-0.151$ \, [$-0.185$, $-0.117$] \\
\quad $+$ retrieval & 0.7478 & --- & --- & --- & $-0.135$ \, [$-0.168$, $-0.102$] \\
\quad $+$ backbone & 0.8068 & 0.094 & 73.3\% & 35.0\% & $-0.076$ \, [$-0.101$, $-0.050$] \\
\quad $+$ backbone, add-only & 0.8030 & 0.013 & 94.8\% & 18.8\% & $-0.080$ \, [$-0.102$, $-0.057$] \\
\midrule
Backbone alone & 0.8484 & --- & --- & --- & $-0.034$ \, [$-0.049$, $-0.021$] \\
Backbone $+$ checked claims (ours) & \textbf{0.8827} & --- & --- & --- & --- (reference) \\
\bottomrule
\end{tabular}
\caption{Frontier LLMs used in place of the deterministic merge, on the same
expert claims and the same evaluation protocol
(\S\ref{method:select}). Every interval excludes zero. Policy compliance
columns are the fraction of backbone diagnoses dropped (drop), the share of
backbone diagnoses whose laterality is preserved (lat.), and the share of
expert claims outside the backbone's coverage that are admitted (admit).
Free-text merger outputs are mapped to canonical diagnoses with one fixed
normalizer shared by every merger arm. Merger outputs are scored canon-only
and so incur no attribute penalties. The merger scores are therefore
favorable estimates, and the reported gap to our system is a lower bound.
Reference-only $F_1$, precision, and recall for these systems are in
Supplementary Table~S16.}
\label{tab:arbiter}
\end{table*}

Providing the backbone improves both LLM mergers, but neither reaches the
backbone alone (GPT-5.6: 0.805; Sonnet-4.6: 0.807; backbone: 0.848; ours:
0.883). An add-only instruction also fails to close the gap. It improves
policy compliance but changes $F_1$ by at most 0.004, and it moves the share
of admitted out-of-backbone claims in opposite directions for the two models
(GPT-5.6: 22\% to 47\%; Sonnet-4.6: 35\% to 19\%; Table~\ref{tab:arbiter}).

\paragraph{Error and per-diagnosis analysis.}
\begin{table}[t]
\centering
\small
\begin{tabular}{@{}lrrrr@{}}
\toprule
System & TP & FP & FN & DIS \\
\midrule
Backbone alone & 414 & 32 & 108 & 50 \\
Ours           & 452 & 39 & \phantom{0}75 & 89 \\
\midrule
$\Delta$ (ours $-$ backbone) & $+38$ & $+7$ & $-33$ & $+39$ \\
\bottomrule
\end{tabular}
\caption{Error counts for the backbone and our selective combination. DIS denotes a prediction absent from the physician reference but meeting the predefined evidence criteria.}
\label{tab:errors}
\end{table}
The improvement from selective augmentation is primarily a recall gain.
Relative to the backbone, our system recovers 33 reference diagnoses that
would otherwise be missed, reducing FNs from 108 to 75, while FPs increase
from 32 to 39 (Table~\ref{tab:errors}). In raw counts, the share of reference
diagnoses found rises from 79.3\% (414/522) to 85.8\% (452/527), and
the share of predictions that match the reference falls from 83.5\%
(414/496) to 77.9\% (452/580). Measurement-supported disagreements
(DIS) increase from 50 to 89, consistent with the smaller gain under
reference-only $F_1$ reported above. Thus, the higher primary $F_1$ reflects
fewer missed diagnoses at the cost of more additional predictions, many of
which satisfy the predefined measurement-evidence criteria. The gains are
concentrated in diagnoses where the backbone is weak, the most frequent
diagnoses change little, and no diagnosis decreases (Supplementary
Table~S9).

\section{Discussion}

Our main finding is that disease-specific experts are more useful as supplements
to a domain-adapted model than as replacements for it. No merging method
exceeded the backbone after it was discarded, whereas adding checked expert
claims mainly recovered diagnoses on which the backbone had lower recall.
Because each expert already received the backbone prediction, the expert-only
failure cannot be attributed to lack of access to it. Instead, retaining the
backbone appears to preserve information that is not reliably recovered after
decomposition, possibly because training on complete studies captures
relationships across diagnostic categories
\citep{read2011classifierchains}. This favors using specialization to correct
weaknesses of a general model rather than partitioning the full diagnostic task.

How expert claims are combined also matters. Unchecked claims did not
significantly improve on the backbone, and frontier-LLM mergers remained below
the deterministic merge even when given the backbone prediction and an
add-only instruction. The deterministic approach applies the same predefined
checks to every claim, avoiding variation in how LLMs follow the combination
instruction \citep{sclar2024promptformat}.

Stage~2 supports separating diagnostic prediction from report generation.
Providing predicted diagnostic slots improved diagnosis-level $F_1$ in the
generated reports for both generators, with further gains from reference slots.
Thus, an explicit diagnostic state can guide report generation while keeping
diagnostic prediction and text generation separately testable
\citep{guo2024dictllm,wei2022cot,nye2021scratchpad}.

The intended use is physician-reviewed draft reporting. Because the model does
not receive patient history, symptoms, or physical examination findings, it cannot
reproduce decisions that depend on them. Our selectively augmented model
recovers diagnoses missed by the backbone, but also introduces more predictions
absent from the physician report. A prospective reader study is needed to determine whether this
improves physician-reviewed reporting \citep{gaube2021aiadvice}.

\section*{Limitations}
All data come from one institution and one language, limiting generalization
across hospitals. The model does not receive symptoms or physical examination
findings. Expert domains were selected on a separate 694-study set and evaluated on a
single held-out 404-study set; however, both sets originate from the same
institutional cohort. The evidence criteria that exempt DIS from the primary
$F_1$ are similar to the checks that admit expert claims; reference-only
$F_1$ does not use these criteria. Report quality was judged by
frontier models rather than physicians, and physician references may
themselves vary between readers \citep{johnsen1995variation}.

\section{Conclusion}

Disease-specific experts helped when they supplemented, rather than replaced, a
domain-adapted backbone, particularly by recovering diagnoses the backbone
missed. Structured diagnostic predictions also improved downstream report
generation. Together, these results favor retaining a domain-adapted generalist,
using experts selectively for its weaknesses, and separating diagnostic
prediction from report generation.

\section*{Ethical Statement}
This study uses de-identified patient records under institutional review board
approval. Direct identifiers were removed before modeling. External model APIs
received de-identified EMG/NCS measurements, model predictions, reference
diagnostic slots (Stage-2 condition~B), and de-identified physician report
text used as the judging reference; no direct patient identifiers were transmitted. The proposed system
produces draft interpretations for physician review and was not evaluated for
autonomous clinical use.

\bibliography{refs}

\setcounter{section}{0}
\setcounter{table}{0}
\setcounter{figure}{0}
\renewcommand{\thesection}{S\arabic{section}}
\renewcommand{\thetable}{S\arabic{table}}
\renewcommand{\thefigure}{S\arabic{figure}}

\section*{Supplementary Material}
\noindent
Section and table numbers prefixed with S refer to this supplement;
unprefixed numbers refer to the main paper.

\section{Laterality Normalization}
A bilateral slot is expanded into separate left and right slots only when
the system it is compared against explicitly reports a lateralized slot for the
same diagnosis; slots with unspecified laterality are unchanged. The rule is
applied identically to every system.
Normalization changes the backbone from 0.8045 to 0.8484 and our system from 0.8451 to 0.8827, so the gap we report narrows from $+0.0406$ $[0.025,0.058]$ to $+0.0343$ $[0.021,0.049]$. The rule adds no information; it only makes a bilateral
label and the corresponding left/right pair count as the same finding.
Because the expansion is conditional on what the compared system reports, the
number of scored reference slots is not identical across systems; this is why
the true positives and false negatives in Table~\ref{tab:errors} of the main paper sum to 522
for the backbone and 527 for our system.

\section{Composition of eval404}
\label{sec:cohort}
The evaluation set was drawn from the 15{,}247-study cohort by stratifying on a
single binary flag, whether the study was normal or abnormal. It was not
stratified by diagnosis, and Table~\ref{tab:cohort} gives the consequence:
CTS appears in about two fifths of the studies, four diagnoses
appear in none, and several appear in fewer than five. We report the full
distribution rather than the head of it, because the tail is what bounds the
claims: any statement about neuromuscular junction disorder, motor
polyneuropathy, small-fiber neuropathy, or multifocal motor neuropathy is
outside what this set can support, and per-diagnosis figures for the diagnoses
at the bottom of the table rest on a handful of cases each.
Counts are reference diagnosis labels rather than patients, so a study that
records a diagnosis separately on each side contributes two; this is why CTS
shows 194 labels but 39.4\% of studies (159 of 404). Percentages sum to more
than 100 because a study may carry several diagnoses.

\begin{table}[t]
\centering
\small
\begin{tabular}{lrr}
\toprule
Diagnosis & Labels & \% of set \\
\midrule
Carpal tunnel syndrome           & 194 & 39.4 \\
Normal study                     & 101 & 25.0 \\
Lumbosacral radiculopathy        &  62 & 13.4 \\
Cervical radiculopathy           &  58 & 13.1 \\
Ulnar neuropathy at the elbow    &  51 & 12.1 \\
Sensorimotor polyneuropathy      &  41 & 10.1 \\
Sensory polyneuropathy           &  22 &  5.4 \\
Motor neuron disease             &  14 &  3.5 \\
Myopathy                         &  12 &  3.0 \\
Axonal polyneuropathy            &  10 &  2.5 \\
Focal mononeuropathy             &   9 &  2.2 \\
Plexopathy                       &   5 &  1.2 \\
Demyelinating polyneuropathy     &   4 &  1.0 \\
Ulnar neuropathy, other site     &   4 &  1.0 \\
Other mononeuropathy             &   3 &  0.7 \\
Sensory neuronopathy             &   2 &  0.5 \\
Median neuropathy, other site    &   2 &  0.5 \\
Mononeuropathy multiplex         &   1 &  0.2 \\
\midrule
Neuromuscular junction disorder  &   0 &  0.0 \\
Motor polyneuropathy             &   0 &  0.0 \\
Small-fiber neuropathy           &   0 &  0.0 \\
Multifocal motor neuropathy      &   0 &  0.0 \\
\bottomrule
\end{tabular}
\caption{Full reference diagnosis distribution of the 404-case evaluation set.
The count column gives reference diagnosis labels; the percentage column
gives the share of the 404 studies carrying the diagnosis at least once.
The set was stratified on a binary normal/abnormal flag only, not by diagnosis.
As a result neuromuscular junction disorder, motor polyneuropathy, small-fiber
neuropathy, and multifocal motor neuropathy have zero cases and cannot be
judged by this evaluation; every
other diagnosis is included, however small its sample.}
\label{tab:cohort}
\end{table}

\section{Judge Rubric and Agreement}

\begin{table*}[t]
\centering
\begin{tabular}{lcccc}
\toprule
System & Correctness & Completeness & Hallucination & Omission \\
\midrule
Reference slots & 4.89 / 4.85 & 4.91 / 4.92 & 0.07 / 0.08 & 0.05 / 0.04 \\
\midrule
Backbone alone & \textbf{4.66 / 4.63} & 4.57 / 4.58 & \textbf{0.19 / 0.21} & 0.24 / 0.23 \\
Ours & 4.51 / 4.47 & \textbf{4.75 / 4.74} & 0.29 / 0.31 & \textbf{0.14 / 0.14} \\
\midrule
GPT-5.6 merger $+$ retrieval (no backbone) & 4.46 / 4.44 & 4.21 / 4.29 & 0.30 / 0.32 & 0.68 / 0.52 \\
Sonnet-4.6 merger $+$ retrieval (no backbone) & 4.35 / 4.34 & 4.21 / 4.27 & 0.38 / 0.39 & 0.68 / 0.52 \\
\bottomrule
\end{tabular}
\caption{Clinical evaluation of Stage-1 predictions by two blinded LLM judges.
Each cell reports Gemini-3.6-flash / DeepSeek-v4-flash. The LLM merger rows are
the retrieval arms without the backbone. Correctness and
completeness are scored from 1 to 5; hallucination and omission are mean counts
per report (lower is better). Evaluation details are given in
Section~\ref{sec:metrics} of the main paper.}
\label{tab:judge}
\end{table*}

\begin{table}[t]
\centering
\small
\begin{tabular}{@{}lcccc@{}}
\toprule
& \multicolumn{2}{c}{GPT-5.6} & \multicolumn{2}{c}{Sonnet-4.6} \\
\cmidrule(lr){2-3}\cmidrule(lr){4-5}
Input & corr. & compl. & corr. & compl. \\
\midrule
A & 4.47/4.40 & 4.74/4.71 & 4.03/3.99 & 4.49/4.47 \\
B & 4.88/4.85 & 4.88/4.88 & 4.87/4.84 & 4.91/4.88 \\
\textbf{C (ours)}
  & \textbf{4.50/4.45} & \textbf{4.75/4.75} & \textbf{4.49/4.40} & \textbf{4.75/4.76} \\
D & 4.50/4.46 & 4.70/4.70 & 4.50/4.40 & 4.74/4.72 \\
\bottomrule
\end{tabular}
\caption{LLM evaluation of generated reports. A: raw measurements only;
B: raw measurements with reference slots; C: raw measurements with predicted
slots; D: predicted slots only. Scores are correctness (corr.) and completeness
(compl.) on a 1--5 scale, reported as Gemini-3.6-flash / DeepSeek-v4-flash.}
\label{tab:stage2judge}
\end{table}

\begin{table*}[t]
\centering
\small
\begin{tabular}{lcc}
\toprule
Comparison & $\Delta$ Compl. & $\Delta$ Corr. \\
\midrule
vs.\ backbone        & $+0.171^{*}$ / $+0.158^{*}$ & $-0.153^{*}$ / $-0.153^{*}$ \\
vs.\ GPT-5.6 merger     & $+0.535^{*}$ / $+0.450^{*}$ & $+0.045$ / $+0.035$ \\
vs.\ Sonnet-4.6 merger  & $+0.535^{*}$ / $+0.470^{*}$ & $+0.158^{*}$ / $+0.129^{*}$ \\
vs.\ reference       & $-0.166^{*}$ / $-0.188^{*}$ & $-0.381^{*}$ / $-0.376^{*}$ \\
\bottomrule
\end{tabular}
\caption{Paired differences (ours $-$ comparison) for the judge scores in
Table~\ref{tab:judge}, by case-level paired bootstrap (404 cases, 10{,}000
resamples, seed 20260724); Gemini / DeepSeek.
$^{*}$ marks $p<0.05$, and most entries are $p<0.0001$.
These are the only $p$-values reported in this paper; every other
comparison is given as a bootstrap interval.
Ours is significantly more complete than every
comparison system under both judges, at the cost of a small correctness
decrement relative to the fine-tuned baseline. Judge reliability:
self-consistency weighted $\kappa$ 0.87--0.95; inter-judge
$\kappa$ 0.93 for correctness and 0.84 for completeness; the two judges agree
in sign on all eight comparisons.}
\label{tab:judgedelta}
\end{table*}

Two models score every report: Gemini-3.6-flash and DeepSeek-v4-flash, both at
temperature zero, both blind to which system produced which text, with
presentation order randomized per case. Neither is among the systems being
compared, so neither can prefer its own output. The reference shown to the judge
is the attending physician's report with sentences that rule a diagnosis
\emph{out} removed, since a system is not expected to reproduce an exclusion it
was never asked to make.
The candidate text is not the system's raw output: each system's diagnostic
slots are rendered into short clinical sentences with a fixed dictionary, so
every system is judged in the same prose form. Parsing those sentences back
recovers the original slots in every case, so the rendering neither adds nor
loses diagnostic information.

\paragraph{What the judge is asked.}
The unit of judgment is the \emph{diagnosis}, not the sentence. Physicians
support one diagnosis with several observations (involved muscles, amplitudes,
the distribution of denervation), and counting each unmatched observation as a
separate error collapses every system toward the same score. The judge is
therefore asked to enumerate the diagnoses in the reference, the diagnoses in
the candidate, and to align them before scoring. It is instructed not to
penalize disagreement about severity, for which competing grading schemes exist
\citep{bland2000grading,robinson2008grading}, and not to reward or penalize
length.

\paragraph{The four axes.}
\emph{Correctness} and \emph{completeness} are scored 1--5 and reported
separately rather than averaged, because our system is expected to move them in
opposite directions. \emph{Hallucination} is the count, per report, of diagnoses
asserted by the candidate that the reference does not contain and the study does
not support. \emph{Omission} is the count, per report, of reference diagnoses
absent from the candidate. Both are counts per report rather than rates over
diagnoses, so a study with four diagnoses can contribute more than one of each.

\paragraph{The validation row.}
Because a rubric can be wrong in ways that no system comparison would reveal, we
score the reference slots themselves, rendered through the same fixed dictionary
as every system. They reach 4.89 and 4.85 for correctness and 4.91 and 4.92 for
completeness on the two judges. A rendering, rubric, or judge that could not
recognize the physician's own findings would fail here, and we treat a score
below 4.5 on this row as a gate.

\paragraph{Agreement.}
For the per-system judging in Table~\ref{tab:judge}, inter-judge weighted
$\kappa$ is 0.93 for correctness and 0.84 for completeness, and repeated scoring
of the same reports by the same judge gives weighted $\kappa$ between 0.87 and
0.95. For the per-condition judging in Table~\ref{tab:stage2judge}, inter-judge
weighted $\kappa$ is 0.872 for correctness and 0.837 for completeness over
3{,}636 shared judgments. The two judges agree in sign on all four
completeness comparisons in Table~\ref{tab:judgedelta}, under both judges.

\paragraph{Verbatim judge prompt.}
The prompt is reproduced in full in Section~\ref{sec:judgeprompt}.

\section{Why Exempting DIS Is Not Double Leniency}

The concern a partial-credit rule invites is that a system could hedge by emitting a
vague or adjacent label and escaping the penalty for it. Two properties prevent
that here.

\paragraph{The exemption is earned per patient, not granted per label.}
A prediction is exempted only when \emph{that patient's} measurements support
it. A diagnosis that the study does not support is an ordinary false positive
however plausible its label, and one the physician explicitly ruled out is
charged at 1.5 times the weight of an ordinary error under every metric. A
hedging strategy would have to produce, for each individual study, a diagnosis
the measurements support, which is not hedging but diagnosis.

\paragraph{Case A: sensory versus sensorimotor polyneuropathy.}
These labels differ in whether motor fibers are involved, and the raw study
records motor responses directly. A system that predicts sensorimotor where the
reference records sensory has asserted something checkable, and the gate
described in Section~\ref{app:scoring} checks it against the motor studies
before the match is allowed. The reverse direction, predicting sensory where
the reference records sensorimotor, drops an established finding, and no
evidence check can license it; that direction is always a full error. The
asymmetry means the partial credit is available only in the direction that can
be verified, which is the opposite of what a hedging strategy would need.

\paragraph{Case B: radiculopathy subdivision.}
Cervical and lumbosacral radiculopathies are separate families, so a
cervical-for-lumbosacral prediction is a full error. Root-level disagreement is
not handled by the exemption at all: it is handled by the attribute penalty,
which scales credit by the overlap between predicted and reference root sets. A
prediction naming every root would score near zero overlap on most cases rather
than collecting partial credit, so breadth is penalized rather than rewarded.

\paragraph{What remains.}
The decision not to penalize a DIS prediction is ours, and reference-only scoring
exists precisely so that a reader who rejects that value can read the table
without it. We report both, and Table~\ref{tab:judgedelta} adds a rubric that
contains no DIS concept at all.

\section{Expert Training and Serving}

Every trained component in this paper is a QLoRA adapter
\citep{dettmers2023qlora} over one frozen copy of Qwen3-8B \citep{qwen3}: the
backbone, and each of the six experts. The merge stage trains nothing.

\paragraph{Adapter configuration (identical for all components).}
The base is loaded in 4-bit NF4 with double quantization and bfloat16 compute
dtype. The adapter uses rank $r=16$, $\alpha=32$, dropout 0.05, no bias term,
and targets all seven projections (\texttt{q\_proj}, \texttt{k\_proj},
\texttt{v\_proj}, \texttt{o\_proj}, \texttt{gate\_proj}, \texttt{up\_proj},
\texttt{down\_proj}), so attention and feed-forward blocks are both adapted.
This makes 43{,}646{,}976 of 8{,}234{,}382{,}336 parameters trainable, or
0.53\%.

\paragraph{Optimization.}
AdamW over the adapter parameters only, learning rate $2\times10^{-4}$, weight
decay 0. The schedule warms up linearly over the first 3\% of steps and then
decays linearly to a floor of 0.05 of the peak rate. Three epochs, per-device
batch size 1 with gradient accumulation 16 (effective batch 16), bfloat16 with
gradient checkpointing. An adapter snapshot is written at every epoch and every
100 steps. Training used no fixed random seed and the data loader shuffled
without one, so each reported model comes from a single unseeded training run
and retraining would not reproduce the weights. The principal comparisons hold
the trained artifacts fixed and vary only the merge. Bootstrap resampling uses
a fixed seed, stated in the main text.

\paragraph{Sequence budget and train/serve contract.}
Prompts are capped at 4096 tokens and targets at 512, the same budget used at
inference. An example that exceeds either budget raises rather than being
silently truncated, so no training example differs from what the model would see
at serving time. On the polyneuropathy expert, whose prompts are the longest of
the six, the prompt length is 1{,}827 tokens at the median, 2{,}624 at the 90th
percentile,
and 3{,}443 at maximum, and no example of 500 sampled was truncated. Loss is
computed on the target span only. A contract file recording the pipeline
version, sequence budgets, and base model identity is written beside each
adapter and checked at load time.

\paragraph{The six experts.}
Table~\ref{tab:experts} lists the domains and training set sizes. Each
expert receives the backbone's prediction plus the study evidence and
returns a JSON claim for its own domain. The claim carries the evidence the
expert cites for its own decision, but that evidence is re-derived from the
measurements during checking rather than trusted as reported, so an expert
cannot admit a diagnosis by asserting support it does not have.

\begin{table}[t]
\centering
\small
\begin{tabular}{@{}p{2.3cm} p{3.7cm} r@{}}
\toprule
Expert & Domain & Rows \\
\midrule
Polyneuropathy & Presence, axonal vs.\ demyelinating, fiber type & 3{,}919 \\
\addlinespace
Entrapment neuropathy & Focal entrapments, judged independently and per side & 8{,}335 \\
\addlinespace
Needle EMG findings & Radiculopathy by myotomal distribution; myopathy signals & 6{,}144 \\
\addlinespace
Motor neuron disease & With El Escorial category & 774 \\
\addlinespace
Mononeuropathy, plexopathy & Single-nerve territory vs.\ plexus, with adjacent-root checks & 1{,}752 \\
\addlinespace
Neuromuscular junction disorder & Myasthenia gravis only & 231 \\
\bottomrule
\end{tabular}
\caption{The six experts. Entrapment neuropathy, polyneuropathy, needle EMG findings, and motor neuron disease are used by the reported system; the other two are trained but not selected (Section~\ref{sec:selection}). Together the four cover ten canonical diagnoses: CTS and ulnar neuropathy at the elbow; sensory, sensorimotor, axonal, and demyelinating polyneuropathy; myopathy, cervical radiculopathy, and lumbosacral radiculopathy; and motor neuron disease. The \emph{Rows} column gives the
number of training examples used for each expert adapter. Every expert
receives the backbone's prediction as part of its input. Every expert is trained on the training split only, and no expert's training data
include a patient from either evaluation set. The counts overlap across experts. Six polyneuropathy examples exceeding the
4{,}096-token budget were dropped.}
\label{tab:experts}
\end{table}

\paragraph{Model identifiers.}
The API-served models are \texttt{gpt-5.6} (raw-direct baseline and merger),
\texttt{claude-sonnet-4-6} (merger and Stage-2 generator), and
\texttt{gemini-3.6-flash} with \texttt{deepseek-v4-flash} (the two judges).
These are the strings recorded in the call logs.

\paragraph{System prompts.}
The backbone is served with the single-line system prompt
\texttt{"You are an expert electrodiagnostic (EMG/NCS) physician."}
The six experts do not share a template: each carries its own domain prompt
stating the decisions it must make and the rules it must apply. The
polyneuropathy expert, whose prompts are the longest, is reproduced below; the
other
five, together with the merger, judge, and Stage-2 generator prompts, will be
released with the code. Two mechanical changes were made
for typesetting: the Unicode symbols $\le$, $\ge$ and $\rightarrow$ are written
here as \texttt{<=}, \texttt{>=} and \texttt{->}, and internal rule-version
tags of the form \texttt{[Fix 6.1]} are removed. The clinical content is
unchanged.

\begin{lstlisting}
You are an electrodiagnostic specialist in peripheral neuropathies. Given nerve conduction measurements, make these decisions:

1. Is a generalized polyneuropathy (PN) present?
   - A PN can be early/mild with only 1-2 nerves - limited involvement is NOT by itself a reason to exclude PN, especially if velocity is slowed.
   - Do NOT diagnose PN if abnormalities are confined to a single entrapment site.
   - An absent response (e.g., absent sural) can itself be abnormal evidence.

2. Fiber involvement (this CAN have more than one plausible answer - list as ranked candidates): sensory (sensory abnormal, motor normal), sensorimotor (both abnormal), or motor. Rank by fit with a likelihood (high/medium/low) and a reason.

Do NOT judge whether the neuropathy is axonal or demyelinating - that is determined separately. Focus only on whether PN is present and on the fiber involvement.
- Demyelination strictness: A demyelinating call requires motor CV <= cutoff (arm 35, leg 28) in >=2 nerves AND CMAP amp >= 1mV. OR conduction block by area / duration criteria at a SINGLE segment. Segmental amp differences from repeated measurements do NOT constitute conduction block.
- Systemic PN force-fire: IF sensory abnormality in >=2 nerves + bilateral + distal-predominant, OR motor abnormality in >=2 nerves + bilateral + distal-predominant -> fire sensorimotor_pn or axonal_pn even if focal CTS/UNE also present. Focal entrapment does NOT suppress a real systemic PN.
- sensory_pn attribution: sensory_pn requires BILATERAL + >=2 nerves. Unilateral mild sensory abnormality -> focal, NOT systemic. If CTS/UNE fires on nerve X, EXCLUDE that nerve's sensory abnormality from the sensory_pn evidence (no double-counting).
- sensorimotor requires sensory: sensorimotor_pn requires sensory abn >=1 AND motor abn >=1. Motor-only with sensory 0 is NOT sensorimotor - flag it as a motor-only pattern (schema gap).
- Diabetic / metabolic hedge: A length-dependent axonal sensorimotor pattern (regardless of age) is suggestive of a metabolic cause (e.g., diabetes) - clinical correlation needed. Very young (<20) -> also mention hereditary. Use NO hard age cutoff.

Output JSON only:
{"present": true/false,
 "candidates": [{"fiber": "sensory"/"motor"/"sensorimotor", "likelihood": "high"/"medium"/"low", "because": "..."}, ...],
 "ruled_out": [{"fiber": "...", "because": "..."}, ...],
 "confidence_present": 0.0-1.0}
\end{lstlisting}

\paragraph{Serving.}
Given fixed trained adapters, repeated runs on the same input are identical:
at inference the 4-bit base is loaded with the relevant adapter and decoded
greedily (no sampling, single beam, at most 512 new tokens). The expert system prompt is supplied in the
system role at both training and inference, which avoids chat-template skew.
This determinism covers the local pipeline only. The API-served
systems (the LLM mergers, the two judges, and the Stage-2 generators) are
reported from a single call at temperature 0 and are not guaranteed to be
bit-reproducible.

\paragraph{Merge.}
The merge stage is deterministic Python over the JSON claims and the measurement
table, with no model and no GPU. It assembles certain claims into slots; applies
false-positive gates that require localizing measurement evidence before an
entrapment claim is accepted, and that suppress implausible demyelination and
root levels; holds genuinely ambiguous conflicts as unresolved rather than
forcing a threshold; adds focal mononeuropathy slots where no systemic
polyneuropathy is present; and performs fiber attribution, downgrading
sensorimotor to sensory polyneuropathy when the motor abnormality is fully
explained by a coexisting diagnosis. The attribution rule fires in 3 of the
404 evaluation cases. Every slot carries the rule that produced it.

\section{Cost, Latency, and Reproducibility}

Experiments ran on a single NVIDIA RTX A5000 (24\,GB, 230\,W) under CentOS
Stream 8 (Linux 4.18.0), with PyTorch 2.11.0 (CUDA 12.8), transformers 5.10.2,
peft 0.19.1, and bitsandbytes 0.49.2.
All timings are wall-clock from the inference logs of real 404-case runs on
that GPU per expert, greedy decoding, at most
512 new tokens. Figures marked as estimates could not be measured directly and
are labeled where they appear.

\paragraph{Latency.}
Per case, the experts take 23.1\,s (entrapment neuropathy), 18.8\,s
(needle EMG findings), 14.6\,s (mononeuropathy, plexopathy), 10.2\,s
(motor neuron disease), and 8.1\,s (neuromuscular junction disorder); the
polyneuropathy run did not complete under logging and
is estimated at about 20\,s from its model class and decode budget. Running all
six serially on one GPU is about 95 GPU-seconds per case. The backbone
prediction adds an estimated 20--30 GPU-seconds (one inference pass of the
same model class and decode budget), and the merge takes under one second on
CPU.
Total compute is therefore about 115--125 GPU-seconds, or 0.033 GPU-hours, per
case: roughly two minutes of wall-clock on one GPU, or about 50 seconds if the
experts are fanned across four. That figure covers all six experts and
so upper-bounds the reported system, which runs four: entrapment neuropathy, needle EMG findings, polyneuropathy, and motor neuron disease together take about 72 GPU-seconds, giving roughly 92--102 GPU-seconds per case with the backbone included. We quote
the six-expert figure in the cost estimates below because it also covers the
ablation configurations. Training the six expert adapters cost 10--15 GPU-hours once, from about
0.3 GPU-hours for the smallest expert to about 3 for the largest. This figure
covers the expert adapters only; the backbone adapter was trained before
timing was logged and is not included.

\paragraph{Local cost.}
Table~\ref{tab:cost} summarizes what follows. At 0.033 GPU-hours per case,
230\,W, a datacenter overhead factor of 1.5, and
\$0.15 per kWh, electricity is about \$0.0017 per case. Amortizing a \$2{,}200
GPU over three years at 40\% duty adds about \$0.007. A rental-price
cross-check at \$0.35 per GPU-hour, which bundles power and hardware, gives
\$0.012. We therefore quote \textbf{about \$0.01 per case all-in}, of which only
the \$0.002 of electricity is marginal once training is amortized.

\paragraph{API cost.}
The raw-direct baseline is a single call with an input of about 2{,}000 tokens
and an output budget of 800. \emph{Pricing assumption:} official rates for the
frontier model we used were not available to us, so we apply a published
same-generation rate of \$1.25 per million input tokens and \$10 per million
output tokens; substituting the official rate rescales the figures below.
A single call costs \$0.0065 at 400 output tokens and \$0.0105 at the full 800.
If hidden reasoning tokens are billed as output, a case with roughly 1{,}500
hidden and 400 visible tokens costs about \$0.0215. The raw-direct baseline is
therefore \$0.007--0.022 per case.

\begin{table}[t]
\centering
\small
\begin{tabular}{@{}p{2.5cm} p{2.3cm} p{2.3cm}@{}}
\toprule
 & Local (ours) & Frontier API \\
\midrule
Per-case compute & 92--102 GPU-s (0.027 GPU-h) & one call, or seven for a matched decomposition \\
\addlinespace
Per-case cost & $\approx$\$0.01 all-in; \$0.002 marginal & \$0.007--0.022 raw-direct; \$0.05--0.15 matched \\
\addlinespace
One-time cost & 10--15 GPU-h, six expert adapters & none \\
\addlinespace
Inference determinism & greedy decoding, deterministic merge & versioned endpoint, not reproducible at temperature zero \\
\addlinespace
Study text leaves the institution & no & yes \\
\bottomrule
\end{tabular}
\caption{Cost and reproducibility, Stage-1 only. Dollar figures for the API use
a published same-generation rate, stated in the text; substituting the official
rate rescales that column. Compute is quoted for the four experts the reported system runs; all six would be 115--125 GPU-seconds.}
\label{tab:cost}
\end{table}

\paragraph{The comparison that matters.}
On a single call the two are the same order of magnitude. But the raw-direct
baseline is also the weakest configuration in Table~\ref{tab:main} of the main paper;
matching our decomposition through an API would take about seven calls per case,
or \$0.05--0.15, every case, forever, against a one-time 10--15 GPU-hours here.
Three differences are structural rather than monetary. Reproducibility: our base
weights are public, our adapters are releasable, decoding is greedy, and the
merge is deterministic, so a rerun from the released adapters reproduces the
output exactly, whereas a
hosted model is versioned, updated without notice, and not reproducible even at
temperature zero. Releasability: the adapters can be published; hosted weights
cannot. Privacy: the study text is protected health information; in the local
Stage-1 system it never leaves the institution, whereas the API baseline
requires transmitting it to a third party under an appropriate agreement. The
Stage-2 generators we evaluate are external, so this advantage currently covers
the diagnostic stage only. For a clinical
system that must be auditable, these favor the local configuration even where a
single API call looks inexpensive.

\section{Additional Stage-2 Observations}

This section collects observations from the Stage-2 runs that the main text
refers to but does not have room to report.
First, a
Qwen3-8B generator shows no significant gain from reference slots over raw
tables, because it follows the slot instruction on only 37\% of cases and
otherwise ignores or embellishes it; we therefore report Stage-2 with the two
frontier generators and note that excluding the generator with the largest raw
slot effect makes our reported effect conservative. Second, across the
slot-conditioned conditions with predicted slots the generated reports reproduce
between 83\% and 97\% of the canonical diagnoses present in the input slots
(94\% and 97\% with raw measurements, 83\% and 91\% without, for GPT-5.6 and
Sonnet-4.6), so the rendering step is close
to lossless and the propagation loss we report is dominated by prediction error
rather than by generation. Third, the few-shot examples used in generation
are two carpal-tunnel cases from the training split, and neither appears in
the evaluation set.

\paragraph{Normal and abnormal studies.}
Table~\ref{tab:stage2split} splits the runs by whether the reference records a
diagnosis. Preservation can only be measured where there is something to
preserve, so normal studies are reported instead as the rate at which the
generator calls the study normal. The raw measurements act in opposite
directions on the two halves. On abnormal studies they add recall: removing them
(D versus C) costs 0.093 for GPT-5.6. On normal studies they invite over-call:
the same generator calls a study normal in 94.2\% of cases from raw tables alone
but 88.3\% when predicted slots are added. The net effect over the whole set is
positive, which is why Table~\ref{tab:stage2} of the main paper reports the combined figure.

The clinical judging is diluted at the 404-case level because normal studies sit
at the rubric ceiling under every condition. Even restricted to the 301 abnormal
studies, GPT-5.6's correctness gain from predicted slots is not significant under
either judge ($+0.086$, $[-0.040, +0.213]$; $+0.113$, $[-0.020, +0.246]$), whereas Sonnet-4.6's is ($+0.575$, $[+0.425, +0.728]$; $+0.528$, $[+0.369, +0.688]$).

\begin{table}[t]
\centering
\small
\begin{tabular}{lcccc}
\toprule
& A & B & C & D \\
\midrule
\multicolumn{5}{l}{\emph{Abnormal studies ($n=301$), preservation}} \\
GPT-5.6    & 0.761 & 0.914 & 0.818 & 0.725 \\
Sonnet-4.6 & 0.626 & 0.899 & 0.782 & 0.775 \\
\addlinespace
\multicolumn{5}{l}{\emph{Normal studies ($n=103$), \% called normal}} \\
GPT-5.6    & 94.2 & 95.1 & 88.3 & 94.2 \\
Sonnet-4.6 & 82.5 & 90.3 & 85.4 & 92.2 \\
\addlinespace
\multicolumn{5}{l}{\emph{Normal studies, invented diagnoses (count)}} \\
GPT-5.6    & 6 & 7 & 13 & 6 \\
Sonnet-4.6 & 19 & 11 & 19 & 12 \\
\bottomrule
\end{tabular}
\caption{The same runs split by whether the reference records a diagnosis. The
split follows the reference partition, under which two boundary cases whose only
abnormality is ultrasound nerve enlargement count as normal; the
reference-label distribution in Section~\ref{sec:cohort} counts those two as
abnormal and therefore lists 101 normal studies.}
\label{tab:stage2split}
\end{table}

\paragraph{The round trip has a ceiling.}
Feeding the reference slots through generation and parsing them back returns
0.906, not 1.000, so 0.094 is lost to the narrative stage before any prediction
error enters. No condition exceeds that ceiling, and condition~B measures it.

\paragraph{Slot scores and preservation measure different things.}
Diagnosis preservation is measured after that round trip, so it is not
comparable to the slot scores in Table~\ref{tab:main} of the main paper. Our slots score 0.883, of which 0.799 survives into a generated report: a loss of 0.083 with the raw tables present and 0.164 without them. One measures what the system
concludes, the other what a report retains. As a consistency check between the
two evaluations, condition~A reproduces the raw-direct rows of
Table~\ref{tab:main} of the main paper to within 0.01 (GPT-5.6 0.749 against 0.739); the
residual comes from generating prose and parsing it back rather than emitting
slots directly.

\paragraph{Length does not explain the ordering.}
In both generators the longest condition is A (119 words for GPT-5.6 and 290
for Sonnet-4.6). For Sonnet-4.6 it also has the lowest precision of the four
(0.670); for GPT-5.6 precision is flat across A, C and D (0.896--0.898). The
shortest is D, whose precision is high (0.898 and 0.888) and whose recall is the
weakest for GPT-5.6 (0.600): its shortfall is missing information, not brevity.
Conditioning on predicted slots raises recall for both generators (GPT-5.6 0.642 to 0.722; Sonnet-4.6 0.550 to 0.684) and precision for Sonnet-4.6 (0.670 to 0.849).

\paragraph{Hallucination.}
Supplying predicted slots takes GPT-5.6 from 0.30 to 0.29 invented findings per
report and Sonnet-4.6 from 0.59 to 0.31 (Gemini-3.6-flash, all 404 studies).

\section{Additional Stage-1 Results}

One arm sits off the ladder in Table~\ref{tab:ablation} of the main paper and is reported here
rather than in the main text, because it is not on the path our system takes.
It discards the backbone and operates on the union of the six experts' claims.

\emph{Union $+$ refutation only} applies the first checking stage and stops,
without measurement correction or attribution: 0.8285 becomes 0.8290, a
difference of 0.0005. Our system does not take this branch: it applies all three
checking stages and then adds to the backbone rather than merging without it.
It does not exceed the backbone alone (0.829 against 0.848; $-0.019$,
$[-0.046, +0.007]$). Table~\ref{tab:rag} reports a second
off-ladder variation: supplying retrieved worked examples to the merger.
Reference-only $F_1$, precision, and recall for the eight merger arms of
Table~\ref{tab:arbiter} of the main paper are given in Table~\ref{tab:mergerstrict}, in
Section~\ref{sec:mergerstrict} at the end of this document.

\begin{table}[t]
\centering
\small
\begin{tabular}{@{}lccc@{}}
\toprule
Merger & No retrieval & Retrieval & $\Delta F_1$ \\
\midrule
Qwen3-8B   & 0.6747 & 0.6026 & $-0.0721$ \\
Sonnet-4.6 & 0.7320 & 0.7478 & $+0.0158$ \\
GPT-5.6    & 0.7919 & 0.8026 & $+0.0107$ \\
\bottomrule
\end{tabular}
\caption{Effect of supplying retrieved worked examples to the merger, in $F_1$.
The 8B merger is included as a smaller-capacity comparison and is not part of
any configuration reported in the main text.}
\label{tab:rag}
\end{table}

\paragraph{Per-diagnosis performance on the primary set.}
Table~\ref{tab:percanon} breaks the primary-set comparison down by canonical
diagnosis. The four largest gains fall on diagnoses where the backbone scored
0.824 or lower, while the two most frequent diagnoses barely move; no diagnosis
decreases.

\begin{table*}[t]
\centering
\small
\setlength{\tabcolsep}{4pt}
\begin{tabular}{@{}lrrrrr@{}}
\toprule
Diagnosis & Ref.\ $n$ & Backbone & Ours & $\Delta F_1$ & $\Delta$ R \\
\midrule
Ulnar neuropathy, elbow       &  52 & 0.702 & 0.879 & $+0.177$ & $+0.269$ \\
Axonal polyneuropathy         &  10 & 0.667 & 0.824 & $+0.157$ & $+0.200$ \\
Myopathy                      &  12 & 0.824 & 0.947 & $+0.123$ & $+0.167$ \\
Sensory polyneuropathy        &  22 & 0.789 & 0.894 & $+0.105$ & $+0.273$ \\
\midrule
Carpal tunnel syndrome        & 212 & 0.942 & 0.948 & $+0.006$ & $+0.019$ \\
Cervical radiculopathy        &  60 & 0.928 & 0.949 & $+0.021$ & $+0.034$ \\
Lumbosacral radiculopathy     &  66 & 0.870 & 0.870 & $0.000$  & $0.000$  \\
Sensorimotor polyneuropathy   &  41 & 0.800 & 0.821 & $+0.021$ & $+0.048$ \\
Motor neuron disease          &  14 & 0.963 & 1.000 & $+0.037$ & $+0.071$ \\
\bottomrule
\end{tabular}
\caption{Per-diagnosis $F_1$ on the 404-case set. Reference $n$ counts scored
slots rather than studies: it reflects bilateral-expanded scoring support and
so is at least as large as the label counts in Table~\ref{tab:cohort}. The upper block holds the
four diagnoses with the largest gains; the lower block holds frequent diagnoses that change little, and motor neuron disease, the domain contributed by the fourth expert.}
\label{tab:percanon}
\end{table*}

\section{Held-Out Expert-Selection Set}
\label{sec:selection}
The expert domains admitted by our system were selected on a held-out set drawn
from the same cohort as eval404: 698 studies, of which 694 carry a backbone
prediction and are scored. This set is used for selection only. The expert
configuration was fixed on it and then evaluated on eval404, which is the
evaluation set throughout the paper. The scorer, the laterality normalization,
and the bootstrap are identical on both. Table~\ref{tab:external} gives the
scores on each. Per-diagnosis results on eval404 are in
Table~\ref{tab:percanon} of the preceding section.

\begin{table}[t]
\centering
\small
\setlength{\tabcolsep}{4pt}
\begin{tabular}{@{}llccc@{}}
\toprule
Set & System & $F_1$ & P & R \\
\midrule
Evaluation (404) & Backbone alone & 0.8484 & 0.925 & 0.784 \\
 & \quad $+$ checked (ours) & \textbf{0.8827} & 0.917 & 0.851 \\
\addlinespace
Selection (694) & Backbone alone & 0.7820 & 0.862 & 0.716 \\
 & \quad $+$ checked (ours) & \textbf{0.8143} & 0.851 & 0.780 \\
\bottomrule
\end{tabular}
\caption{Backbone and final configuration on each set (patient-disjoint models). $\Delta$ is $+0.0343$ $[+0.021, +0.049]$ on the evaluation set and $+0.0323$ $[+0.0212, +0.0439]$ on the selection set; both intervals exclude zero. The selection-set figures are
given for completeness: that set chose the expert domains, so it is not an
independent measure of performance.}
\label{tab:external}
\end{table}

\paragraph{Where the gain lands.}
Table~\ref{tab:extclass} counts the reference diagnoses each system recovers on
the selection set. The gain is confined to the ten diagnoses our system is allowed to add, and within those it is concentrated where the backbone misses most:
ulnar neuropathy at the elbow alone accounts for seventeen recovered cases, its
misses falling from 27 to 10. The recall gain has a matching cost on normal
studies, where the same additions turn 24 correct empty predictions into
incorrect ones. The four diagnoses that appear only in this set
lie outside those ten, and our system leaves them exactly as the backbone left them. This is also why this set says nothing about them.

\begin{table}[t]
\centering
\small
\begin{tabular}{@{}lcc@{}}
\toprule
Diagnosis & Backbone & Ours \\
\midrule
\multicolumn{3}{@{}l}{\textit{Inside the ten domains we add}} \\
Carpal tunnel syndrome & 217 & 219 \\
Cervical radiculopathy & 77 & 80 \\
Lumbosacral radiculopathy & 75 & 80 \\
Ulnar neuropathy at the elbow & 64 & 81 \\
Sensorimotor polyneuropathy & 46 & 51 \\
Sensory polyneuropathy & 17 & 27 \\
Motor neuron disease & 13 & 14 \\
Normal study & 181 & 157 \\
\addlinespace
\multicolumn{3}{@{}l}{\textit{Outside them (recovered / present)}} \\
Neuromuscular junction disorder & 4 / 7 & 4 / 7 \\
Motor polyneuropathy & 0 / 2 & 0 / 2 \\
Small-fiber neuropathy & 0 / 2 & 0 / 2 \\
Multifocal motor neuropathy & 0 / 1 & 0 / 1 \\
\bottomrule
\end{tabular}
\caption{Reference diagnoses recovered on the selection set. The four
diagnoses below the rule are identical for both systems by construction: they
fall outside the ten domains, so no expert claim about them is ever added. We
report them as counts rather than $F_1$ because there are at most seven of each.
The normal-study row counts the 198 studies with no reference diagnosis that
the system also leaves empty. It moves in the opposite direction to the rows
above it: because the merge only adds diagnoses, any diagnosis added to a normal
study costs one, and ours loses 24 of them.}
\label{tab:extclass}
\end{table}

\paragraph{Why the absolute scores are lower.}
Both systems score lower here than on eval404, and by about the same amount
(0.066 for the backbone, 0.068 for ours), so the shift is a property of the
draw rather than of either method. Table~\ref{tab:extcomp} shows where it comes
from. The selection set is less concentrated on CTS, which is the
highest-scoring diagnosis in our evaluation, and it carries four diagnoses that
eval404 does not contain at all. The burden of multiple diagnoses per case
is if anything slightly lower, so the drop is a change in class mix and not in
case complexity.

\begin{table}[t]
\centering
\small
\setlength{\tabcolsep}{3pt}
\begin{tabular}{@{}lcc@{}}
\toprule
 & Evaluation (404) & Selection (694) \\
\midrule
Normal studies & 25.0\% & 29.7\% \\
Cases with $\geq 2$ diagnoses & 26.7\% & 24.9\% \\
Diagnoses per abnormal case & 1.47 & 1.48 \\
Distinct diagnoses present & 17 & 21 \\
Carpal tunnel syndrome & 39.4\% & 33.1\% \\
Ulnar neuropathy at the elbow & 12.1\% & 13.1\% \\
Lumbosacral radiculopathy & 13.4\% & 12.8\% \\
Cervical radiculopathy & 13.1\% & 12.1\% \\
Sensorimotor polyneuropathy & 10.1\% & 10.2\% \\
Sensory polyneuropathy & 5.4\% & 5.6\% \\
Focal mononeuropathy & 2.2\% & 4.3\% \\
Motor neuron disease & 3.5\% & 2.3\% \\
Myopathy & 3.0\% & 1.3\% \\
\bottomrule
\end{tabular}
\caption{Composition of the two sets, as a share of cases. Neuromuscular
junction disorder, small-fiber neuropathy, multifocal motor neuropathy and
motor polyneuropathy occur only in the selection set, each in under 1.5\% of cases;
they are too few to support a per-diagnosis figure, and we make no claim about
them.}
\label{tab:extcomp}
\end{table}

\paragraph{Expert-domain selection.}

We selected the expert domains on the 694-case set by scoring all 63 non-empty subsets of the six trained experts and taking the highest-scoring subset. Table~\ref{tab:subset694-size} reports the best subset at each size, and Table~\ref{tab:subset694-all} gives the full ranking of the 15 four-expert subsets. Entrapment, polyneuropathy, needle EMG, and MND score highest, at 0.8143. The margins at the top are small: the best three-expert subset (the same without MND) scores 0.8135 ($-0.0008$, 95\% CI $[-0.0024, 0.0000]$), and the next-best four-expert subset (NMJ in place of MND) scores 0.8125 ($-0.0018$, $[-0.0041, 0.0000]$), because the MND and NMJ domains touch few of the 694 studies. Admitting all six domains scores 0.8078, below the chosen four ($-0.0065$, $[-0.0105, -0.0030]$). This configuration was then
fixed before evaluation on eval404; no expert-domain selection
was performed on eval404.

\begin{table*}[t]
\centering
\small
\begin{tabular}{@{}clcl@{}}
\toprule
$k$ & Best $k$-expert subset & $F_1$ & $\Delta F_1$ vs.\ chosen [95\% CI] \\
\midrule
0 & backbone alone                                   & 0.7820 & $-0.0323$ \, [$-0.0439$, $-0.0212$] \\
1 & entrapment                                       & 0.7948 & $-0.0194$ \, [$-0.0281$, $-0.0110$] \\
2 & entrapment $+$ PN                                & 0.8050 & $-0.0092$ \, [$-0.0146$, $-0.0045$] \\
3 & entrapment $+$ PN $+$ needle                     & 0.8135 & $-0.0008$ \, [$-0.0024$, $\phantom{-}0.0000$] \\
4 & \textbf{entrapment $+$ PN $+$ needle $+$ MND}    & \textbf{0.8143} & --- (chosen) \\
5 & \quad $+$ NMJ                                    & 0.8132 & $-0.0010$ \, [$-0.0027$, $\phantom{-}0.0000$] \\
6 & all six                                          & 0.8078 & $-0.0065$ \, [$-0.0105$, $-0.0030$] \\
\bottomrule
\end{tabular}
\caption{Expert-subset selection on the 694-case set. For each subset size $k$
we report the best of all
$\binom{6}{k}$ subsets, using add-only merge onto the backbone and
bilateral-credit $F_1$. The chosen four are the highest-scoring subset of any size. Dropping the MND expert ($k=3$) or adding the neuromuscular junction expert ($k=5$) changes $F_1$ by at most 0.001, with intervals that reach zero, because those domains touch few of the 694 studies; removing further experts is significantly worse ($k=1$, $k=2$). The drop at $k=6$
is driven by the mononeuropathy-plexopathy expert, the only domain entering
between $k=5$ and $k=6$. Row $k=0$ is the backbone alone on this set. Domains: entrapment (carpal tunnel syndrome, ulnar
neuropathy at the elbow), PN (polyneuropathy), needle (needle EMG), MND (motor
neuron disease), NMJ (neuromuscular junction), plex
(mononeuropathy/plexopathy). Intervals are 95\% paired bootstrap over
10{,}000 resamples.}
\label{tab:subset694-size}
\end{table*}

\begin{table*}[t]
\centering
\small
\begin{tabular}{@{}rlc@{\hspace{2em}}rlc@{}}
\toprule
\# & Four-expert subset & $F_1$ & \# & Four-expert subset & $F_1$ \\
\midrule
\textbf{1} & \textbf{ent $+$ PN $+$ needle $+$ MND} & \textbf{0.8143} & 9  & ent $+$ PN $+$ plex $+$ NMJ     & 0.7985 \\
2  & ent $+$ PN $+$ needle $+$ NMJ   & 0.8125 & 10 & ent $+$ needle $+$ plex $+$ NMJ & 0.7974 \\
3  & ent $+$ PN $+$ needle $+$ plex  & 0.8081 & 11 & PN $+$ needle $+$ MND $+$ plex  & 0.7970 \\
4  & ent $+$ PN $+$ MND $+$ NMJ      & 0.8048 & 12 & PN $+$ needle $+$ plex $+$ NMJ  & 0.7952 \\
5  & ent $+$ needle $+$ MND $+$ NMJ  & 0.8037 & 13 & ent $+$ MND $+$ plex $+$ NMJ    & 0.7891 \\
6  & PN $+$ needle $+$ MND $+$ NMJ   & 0.8015 & 14 & PN $+$ MND $+$ plex $+$ NMJ     & 0.7871 \\
7  & ent $+$ PN $+$ MND $+$ plex     & 0.8003 & 15 & needle $+$ MND $+$ plex $+$ NMJ & 0.7859 \\
8  & ent $+$ needle $+$ MND $+$ plex & 0.7993 &    &                                 &        \\
\bottomrule
\end{tabular}
\caption{All 15 four-expert subsets on the 694-case selection set, ranked by bilateral-credit $F_1$. Abbreviations are as in Table~\ref{tab:subset694-size}. The selected set ranks first, ahead of the next best by $+0.0018$ $F_1$ (95\% CI $[0.0000, +0.0041]$). For reference, the backbone alone scores 0.7820, the best three-expert subset 0.8135, and all six experts together 0.8078.}
\label{tab:subset694-all}
\end{table*}

\section{Scoring Protocol Details}
\label{app:scoring}
This section states the scoring rules exactly as implemented.

\paragraph{Classifying a predicted slot that finds no match.}
Table~\ref{tab:scoring} gives the three-way classification of an unmatched
prediction and the attribute weights applied to a matched slot.

\begin{table}[t]
\centering
\small
\begin{tabular}{@{}p{5.1cm}c@{}}
\toprule
\multicolumn{2}{@{}l}{\textit{(a) A predicted slot that finds no match}} \\
\midrule
The reference explicitly rules the diagnosis out & FP, weight 1.5 \\
The patient's measurements support it & DIS \\
The measurements do not support it & ordinary FP \\
\addlinespace
\multicolumn{2}{@{}l}{\textit{(b) Attribute factors on a matched slot}} \\
\midrule
Laterality, any diagnosis & $\times 0.5$ \\
Laterality unspecified on one side & no penalty \\
Radiculopathy: root sets & $\times(0.3 + 0.7\,J)$ \\
Radiculopathy: severity & $\times 0.8$ \\
Radiculopathy: non-adjacent skip level & $\times 0.7$ \\
Plexopathy region or focal nerve: unspecified or mixed & $\times 0.5$ \\
\quad outright mismatch & no match at all \\
Severity, any other diagnosis & no penalty \\
A missed reference slot tagged as clinically required & $\times 0.5$ \\
\bottomrule
\end{tabular}
\caption{Scoring rules. $J$ is the Jaccard overlap between predicted and
reference root sets; a matched slot starts at 1.0 and is multiplied by every
factor that applies. Unmatched predictions are separated by the evidence in the
individual study, not by the relationship between the two labels.}
\label{tab:scoring}
\end{table}

\paragraph{What counts as supporting evidence.}
The test is diagnosis-specific and is applied to the patient's own
measurements: at least one abnormal sensory conduction study for sensory
polyneuropathy; abnormal sensory \emph{and} motor studies for sensorimotor
polyneuropathy; localizing evidence for CTS and for ulnar
neuropathy at the elbow; needle-EMG abnormality in the muscles served by the
implicated root for cervical and lumbosacral radiculopathy; fasciculation
together with widespread neurogenic change for motor neuron disease; and
myopathic muscle findings for myopathy.

\paragraph{Sub-type matching.}
Axonal and sensorimotor polyneuropathy are treated as equivalent labels, since
one names the mechanism and the other the fiber type and the two co-occur.
A predicted sensorimotor polyneuropathy matches a reference sensory
polyneuropathy only when motor abnormality is present in the study; the reverse
never matches. The asymmetry has one reason: the direction that asserts more can
be checked against the measurements, while the direction that omits an
established finding cannot be excused by any measurement. The equivalence
applies when a prediction is matched to a reference slot; per-diagnosis
reporting in Table~\ref{tab:percanon} still groups by canonical label, so the
two appear as separate rows there.

\onecolumn
\section{Verbatim Judge Prompt}
\label{sec:judgeprompt}
The text below is the prompt given to both judges, reproduced exactly. The
reference and the candidate are substituted for
\texttt{\{CLEAN\_FINAL\_INTERPRETATION\}} and \texttt{\{PREDICTION\}};
all other text is fixed. It is quoted here as an artifact of the evaluation
and is not an instruction to the reader.

\begin{Verbatim}[fontsize=\footnotesize,
  frame=single,
  framesep=8pt,
  breaklines=true,
  breaksymbolleft={},
  breaksymbolright={}]
You are grading an EMG/NCS diagnostic interpretation.

## Reference (attending physician)
{CLEAN_FINAL_INTERPRETATION}

## Candidate
{PREDICTION}

## UNIT OF SCORING -- read this first
Count DIAGNOSES, not findings or measurements.
The physician lists many measurement findings (active
fasciculations, reduced amplitudes, absent responses,
etc.) as EVIDENCE supporting a single diagnosis. These
are NOT separate abnormalities.
Example: "widespread active denervation, fasciculations,
reinnervation, absent median and ulnar motor responses"
supporting motor neuron disease = ONE diagnosis, not four.

## What counts
COUNT:
- Each distinct diagnosis the physician asserts
- Diagnoses labelled "incidental" -- these count normally
- Diagnoses supported by neuromuscular ultrasound --
  ultrasound is part of the study input

DO NOT COUNT:
- Normal findings ("responses are normal",
  "ultrasound is normal")
- Ruled-out diagnoses
  ("There is no electrodiagnostic evidence of X")
- Findings the physician calls "of unclear significance",
  equivocal, or nondiagnostic
- Diagnoses based only on symptoms, history, or external
  imaging (MRI/CT)
- Differences in wording, sentence order, length, or format

## Equivalences -- treat as identical
- "bilateral" == "right + left"
- Radiculopathy level within ONE level (e.g., C7 vs C7-C8)
  is clinically acceptable
- A label more specific than the reference but consistent
  with it (reference "radiculopathy", candidate
  "lumbosacral radiculopathy" when the reference records
  lumbosacral)
- Nerve name vs eponym ("median neuropathy at the wrist"
  == "carpal tunnel syndrome")

## Severity
Severity (mild / moderate / severe) is subjective.
Do NOT lower the score for severity mismatch.

## Scoring -- two independent axes, 1 to 5
CORRECTNESS
(diagnoses asserted that the reference does not support)
5 - none
4 - none, but laterality off on one
3 - one
2 - two
1 - three or more

COMPLETENESS
(diagnoses in the reference that are missing)
5 - none missing
4 - none missing, but laterality off on one
3 - one missing
2 - two missing
1 - three or more missing

## Worked example
Reference: "moderate left median neuropathy at the wrist
(carpal tunnel syndrome) with sensory axon loss, and a
mild left ulnar neuropathy at the elbow. Ultrasound ...
is normal."

Candidate: "left carpal tunnel syndrome; sensory
polyneuropathy; ulnar neuropathy at the elbow"

-> Reference diagnoses = 2 (CTS left, UNE left).
Candidate adds sensory polyneuropathy, unsupported.
correctness 3, completeness 5

## Output -- JSON only, no other text
{"correctness": <1-5>, "completeness": <1-5>,
 "false_findings": ["..."], "missed_findings": ["..."]}
\end{Verbatim}

\twocolumn
\section{Reference-Only Scores for the LLM Mergers}
\label{sec:mergerstrict}
Table~\ref{tab:arbiter} of the main paper reports the merger arms under the primary metric.
Table~\ref{tab:mergerstrict} gives the same arms under reference-only $F_1$,
which treats every prediction absent from the physician reference as an error
and awards no partial credit for attribute agreement. The scorer, the
laterality normalization, and the 404-case evaluation set are those used
throughout. Under this metric the backbone alone scores 0.8134 and our system
0.8166, so both remain above every merger arm, and the ordering between the
deterministic merge and the LLM mergers is unchanged.

\begin{table}[t]
\centering
\small
\begin{tabular}{@{}lccc@{}}
\toprule
System & $F_1$ & P & R \\
\midrule
GPT-5.6 merger                  & 0.6941 & 0.6673 & 0.7232 \\
\quad $+$ retrieval             & 0.7008 & 0.6690 & 0.7358 \\
\quad $+$ backbone              & 0.7260 & 0.7116 & 0.7411 \\
\quad $+$ backbone, add-only    & 0.7357 & 0.6812 & 0.7996 \\
\addlinespace
Sonnet-4.6 merger               & 0.6500 & 0.6081 & 0.6981 \\
\quad $+$ retrieval             & 0.6631 & 0.6281 & 0.7022 \\
\quad $+$ backbone              & 0.7154 & 0.6871 & 0.7461 \\
\quad $+$ backbone, add-only    & 0.7226 & 0.6954 & 0.7520 \\
\bottomrule
\end{tabular}
\caption{Reference-only $F_1$, precision, and recall for the eight merger arms
of Table~\ref{tab:arbiter} of the main paper, on the 404-case evaluation set. Merger outputs
are scored canon-only, as in the main paper. For comparison, the backbone alone
scores 0.8134 and our system 0.8166 under the same metric.}
\label{tab:mergerstrict}
\end{table}

\end{document}